\documentclass[letterpaper, 10 pt, conference]{ieeeconf}  
\IEEEoverridecommandlockouts                              
\usepackage{epsfig} 
\usepackage{times} 
\usepackage{hyperref}
\usepackage{mathtools}

\usepackage{amsmath} 
\usepackage{amssymb}  
\usepackage{mathrsfs}
\usepackage{algorithm,algorithmic}
\usepackage{ amssymb }

\title{\Large \bf
Motion Decoupling and Composition via Reduced Order Model Optimization for Dynamic Humanoid Walking with CLF-QP based Active Force Control
}
    \author{Xiaobin Xiong and Aaron D. Ames
\thanks{*This work is supported by Amazon Fellowship in AI.}
\thanks{The authors are with the Department of Mechanical and Civil Engineering, California Institute of Technology, Pasadena, CA 91125
        {\tt\small xxiong@caltech.edu}, {\tt\small ames@caltech.edu}}%
 }
\begin{document}
\maketitle
\thispagestyle{empty}
\pagestyle{empty}

\begin{abstract}
In this paper, 3D humanoid walking is decoupled into periodic and transitional motion, each of which is decoupled into planar walking in the sagittal and lateral plane. Reduced order models (ROMs), i.e. actuated Spring-loaded Inverted Pendulum (aSLIP) models and Hybrid-Linear Inverted Pendulum (H-LIP) models, are utilized for motion generation on the desired center of mass (COM) dynamics for each type of planar motion. The periodic motion is planned via point foot (underactuated) ROMs for dynamic motion with minimum ankle actuation, while the transitional motion is planned via foot-actuated ROMs for fast and smooth transition. Composition of the planar COM dynamics yields the desired COM dynamics in 3D, which is embedded on the humanoid via control Lyapunov function based Quadratic programs (CLF-QPs). Additionally, the ground reaction force profiles of the aSLIP walking are used as desired references for ground contact forces in the CLF-QPs for smooth domain transitions. The proposed framework is realized on a lower-limb exoskeleton in simulation wherein different walking motions are achieved.
 \end{abstract}
\section{INTRODUCTION}
Inverted pendulums are valuable template models for realizing walking on humanoid robots. Since the inverted pendulum models are lower dimensional, they are often termed {\it reduced order models (ROMs)}. One of the most commonly used ROMs is the Linear Inverted Pendulum (LIP) \cite{kajita2003biped} \cite{pratt2012capturability} that is applied in the Zero Moment Point (ZMP) approaches \cite{tedrake2015closed} \cite{takenaka2009real} \cite{feng20133d}. The constant height of the point mass yields linear dynamics allowing fast online optimization \cite{kim2013quadratic} for gait generation. Closed-form solutions \cite{tedrake2015closed} have even been suggested for gait generation and stabilization. However, the ZMP walking oftentimes is quasi-dynamic to some extent due to the constant center of mass (COM) height.

The ZMP approach \cite{kajita2003biped} can also be viewed as embedding the LIP dynamics into the full dimensional humanoid robot. 
Similar idea has also appeared in the literature \cite{liu2015dynamic} \cite{Garofalo2012WalkingCO} \cite{mordatch2010robust} \cite{wensing2013high} for embedding the canonical Spring-loaded Inverted Pendulum (SLIP) \cite{full1999templates}. For instance, \cite{Garofalo2012WalkingCO} realized planar periodic bipedal walking in simulation by embedding a SLIP walking dynamics on the robot. \cite{wensing2013high} embedded a 3D-SLIP for humanoid running. The SLIP walking is oftentimes considered dynamic since it generates natural oscillation of the COM and human-like ground reaction forces \cite{geyer2006compliant}. 


The difference in the dynamics of the LIP and SLIP results in different approaches to walking gait generation. The linear dynamics of LIP motivates solvable receding horizon optimizations \cite{kajita2003biped} \cite{kim2013quadratic} \cite{takenaka2009real} for continuous planning and stabilization, thus periodic behaviors \cite{xiong2019IrosStepping} are less studied. On the other hand, the nonlinear and underactuated dynamics of the SLIP drives researchers to study periodic orbits and their stabilization \cite{geyer2006compliant} \cite{Garofalo2012WalkingCO}, instead of nonperiodic behaviors or continuous planning methods. 

In this paper, we propose a composition of LIP and SLIP models for walking to leverage the advantages of each ROM. Walking is firstly decoupled into periodic and nonperiodic/transitional motion. 
\begin{figure}[t]
      \centering
      \includegraphics[width= 3.in]{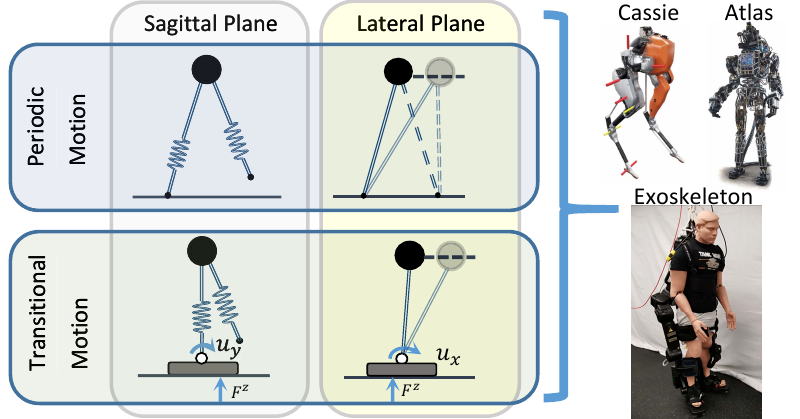}
      \caption{Illustration of the motion decoupling and composition for humanoid walking via reduced order models.}
      \label{overview}
\end{figure}
Both are further decoupled into walking in the sagittal and lateral planes, each of which is characterized by a ROM. Each ROM is modified from the canonical ROMs to best match the type of motion (Fig. \ref{overview}). Specifically, we use an actuated SLIP (aSLIP) model and a Hybrid LIP (H-LIP) model for sagittal and lateral walking, respectively. Point foot models of them are used for synthesizing periodic walking, and the footed models (with ankle actuation) are applied for generating transitional motion. The 3D coupling between planar models comes from the ground normal forces.

To generate optimal walking behaviors, we formulate nonlinear optimization problems on the aSLIP models for the periodic and transitional motion. The periodic orbits of the point foot H-LIP can be directly identified \cite{xiong2019IrosStepping}. The transitional motion of the footed H-LIP is synthesized via a quadratic program (QP). ZMP constraint and ankle actuation limit are included in the transitional optimizations. 
The motion from the ROMs is encoded by the COM, swing foot position and ground reaction forces. We propose an embedding of the desired ROM dynamics on humanoids via control Lyapunov function based Quadratic programs (CLF-QPs). The CLF-QPs formulation also enables force control on matching ground normal forces of the aSLIP on the humanoid, which naturally creates smooth transitions between different domains. 

The composition of ROMs naturally combines the benefit of using each ROM. The aSLIP model enables the walking to be dynamic, and the H-LIP model facilitates solvable QP for planning. The optimization of the aSLIP model is also fast to solve due to its low dimensionality. Moreover, the optimization for each transition is only required for solving once, which potentially can perform online with the recent progresses \cite{ding2019ICRA} \cite{mitCheetah} on the online optimization.
The proposed framework is implemented on an exoskeleton in simulation to generate different walking behaviors, which illustrates periodic motion generation on pointed foot ROMs, the transition optimization on footed ROMs, and finally the CLF-QPs based force control for the dynamics embedding on the robot.
\section{Robot Model}
We use an exoskeleton (Exo) \cite{wander} as the example system for evaluating our methods. Although the exoskeleton is not designed as a humanoid, it does have the identical anthropomorphic leg design to many current humanoids, such as the Atlas robot in Fig. \ref{overview}. Here we briefly describe the robot model and the hybrid model of walking. The hybrid models of the ROMs are described similarly.
\subsection{Dynamics Model}
\label{sec:Modeling}
The Exo has a standard anthropomorphic leg design. Each leg has a hip joint with three degrees of freedom (DOF), a knee joint with one DOF and an ankle joint with two DOF. There are four load cells on each foot for contact force measurement. The dynamics of the system is described with the floating base model by the Euler-Lagrange equation:
\begin{eqnarray}
&& M(q)\ddot{q} + h(q,\dot{q}) = Bu + J_{v}(q)^T F_{v},  \label{eom} \\
&& J_{v}(q) \ddot{q} + \dot{J}_{v}(q) \dot{q} = 0, \label{hol}
\end{eqnarray}
where $q\in SE(3) \times \mathbb{R}^{n = 12}$, $M(q)$ is the mass matrix, $h(q,\dot{q})$ is the Coriolis, centrifugal and gravitational term, $B$ and $u\in \mathbb{R}^{12}$ are the actuation matrix and the motor torque vector, and $F_{v}\in \mathbb{R}^{n_{v}}$ and $J_{v}$ are the holonomic force vector from ground contact and the corresponding Jacobian respectively. We use subscript $v$ to denote different domains. For example, when robot has two feet flatly contacting the ground, i.e. in Double Support Phase (DSP),  $n_{v = \textrm{DSP}} = 12$.

\subsection{Hybrid Model of Two Domain Walking}
In this paper, we are interested in two domain walking, i.e. walking with Single Support Phase (SSP) and Double Support Phase (DSP) depending on the number of legs that contact the ground. We assume that the feet always have flat contact with the ground. The transition from DSP to SSP happens when the one of the stance feet is about to lift off the ground (ground reaction force becomes 0). The transition from SSP to DSP happens when the swing foot strikes the ground. The impact between the swing foot and the ground is modeled as plastic impact \cite{grizzle2014models}, where the velocity of the system undergoes a discrete jump, i.e., 
\begin{align}
& \dot{q}_{\text{DSP}}^{+} = \Delta(q)\dot{q}_{\text{SSP}}^{-},
\end{align}
where $\Delta(q)$ is the impact map. 


%

\section{Periodic Walking Generation via Point Foot Reduced Order Models}
In this section, we describe the periodic motion generation of the point foot reduced order models. Specifically, the actuated Spring-loaded Inverted Pendulum (aSLIP) model is used for generating target COM dynamics in the sagittal plane for periodic walking; the Hybrid passive Linear Inverted Pendulum (H-LIP) model is used for identifying the COM dynamics for the lateral periodic motion. Both models have been introduced in \cite{xiong2018coupling}, and the H-LIP model is further studied in \cite{xiong2019IrosStepping}. Here we briefly describe them for the application of this paper and complement certain details for the ease of understanding.
\subsection{Point Foot aSLIP Model for Sagittal Periodic Walking}
The \textit{point foot} aSLIP model is a SLIP model with leg length actuation (Fig. \ref{PeriodicWalking} (S1)) \cite{xiong2018coupling}. The emphasis on the \textit{point foot} is to differentiate from the later one with foot actuation. SLIP models with actuation have existed in the literature \cite{liu2015dynamic} \cite{poulakakis2009spring} \cite{piovan2015reachability}. The one we use is originally proposed for approximating the walking dynamics of the robot Cassie \cite{xiong2018coupling}. The springs on the aSLIP come from the leg spring approximation of the leg dynamics \cite{xiong2018bipedal}. The main differences of the aSLIP, comparing with the others, are that: the springs have nonlinear stiffness and damping which are functions of leg length $L$; the actuation is realized via changing leg length, i.e. $\ddot{L}$. The dynamics are in the Appendix. 

We use the aSLIP for generating periodic walking motion of the COM. The walking is modeled by two domains, i.e. SSP and DSP (Fig. \ref{PeriodicWalking} (S1)). Specifications of the walking are encoded as equality and inequality constraints in the optimization, such as the durations of each domain, step length or speed and the range of the COM height. Physical constraints include nonnegative contact forces, friction cones, and spring deflection limits.

The optimization is formulated via the direct collocation \cite{hereid20163d} method to avoid integration of the dynamics. The trajectories are discretized over time in each domain with even nodal spacing. We use trapezoidal integration for approximating the dynamics by algebraic constraints. Domain transition constraints are directly enforced at the nodes of the start and end of each domain. With the cost being minimizing the virtual actuation $\ddot{L}$, the optimization is formulated as:
\begin{align}
\label{SLIPopt}
z^* = \underset{z}
 {\text{argmin}}   & \sum_{i = 1}^N \frac{\Delta t}{2} ({{\ddot{L}}_\text{L}}^{i^2} + {{\ddot{L}}_\text{R}}^{i^2}),\\
\text{s.t.}
\quad &  z_{\text{min}} \leq z \leq z_{\text{max}}, \nonumber\\
\quad &  \mathbf{c}_{\text{min}} \leq \mathbf{c}(z) \leq \mathbf{c}_{\text{max}},
\end{align}
where $z$ is the vector of all optimization variables, $\Delta t$ is the time discretization, and $\mathbf{c}(z)$ includes all constraints. The subscripts $\text{L}, \text{R}$ denote the left and right leg. The simple dynamics facilitates fast performance on solving the optimization, which is normally solved within $1 \sim 2$ seconds with random initial guesses using IPOPT \cite{wachter2006implementation}.

The canonical SLIP model can also be used for finding different walking motions \cite{Garofalo2012WalkingCO}. However, the system energy of canonical SLIP is conserved, which is not preferred for generating versatile motions. Thus we use the aSLIP model. The direct use of the stiffness and damping from a physical robot Cassie (Fig. \ref{overview}) also avoids the trial and error on finding the appropriate parameters of the springs.

\subsection{Point Foot H-LIP for Periodic Lateral Motion Generation}
The aSLIP optimization generates periodic motion in the sagittal plane. The durations of each domain, i.e. $T_{\textrm{SSP}}$ and $T_{\textrm{DSP}}$, are thus fixed. Here we apply the Hybrid passive Linear Inverted Pendulum (H-LIP) for generating periodic lateral motion given the domain durations.

\begin{figure}[t]
      \centering
      \includegraphics[width= 3in]{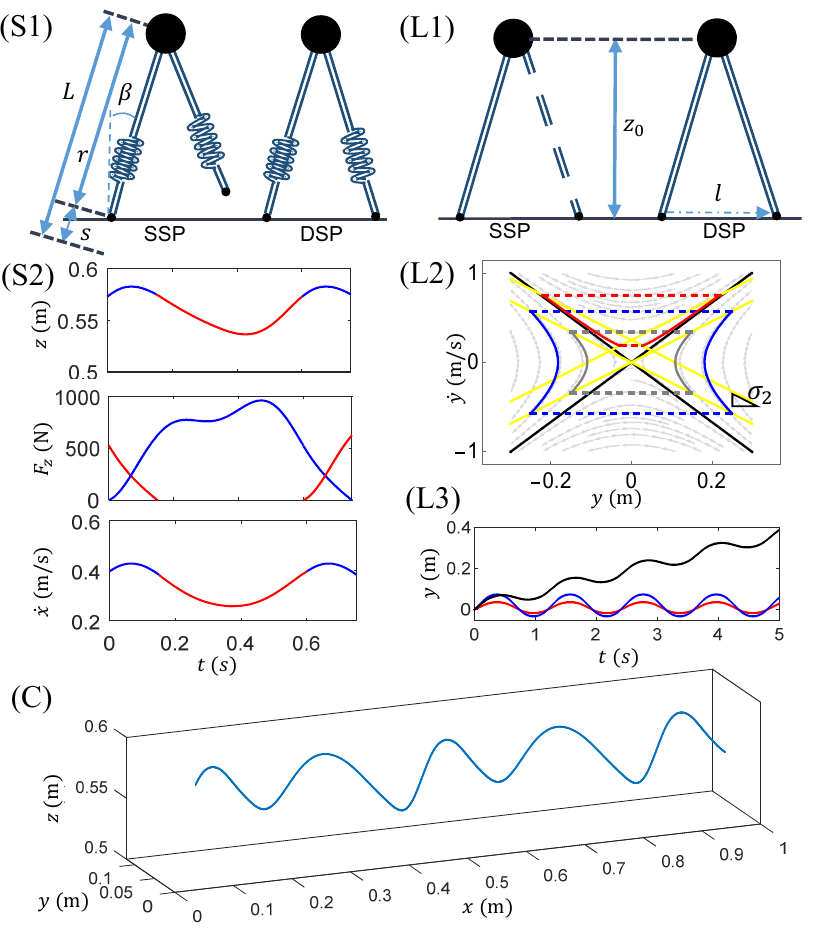}
      \caption{(S1, L2) Illustration of the point foot aSLIP and H-LIP. (S2) Trajectories of an optimized periodic gait in terms of the vertical position $z$, normal reaction force $F_z$ of foot contact and forward velocity $\dot{x}$ of the point mass. (L2) The characterized periodic orbits of the H-LIP in the phase portrait. (L3) Trajectories of the point mass for different periodic orbits. (C) An example of the composed trajectory of the point mass in 3D.} 
      \label{PeriodicWalking}
\end{figure}

The H-LIP is the point foot (passive) Linear Inverted Pendulum (LIP). It also has double support phase (DSP) in walking (Fig. \ref{PeriodicWalking} (L1)). The velocity in DSP is assumed to be constant. We used it in \cite{xiong2018coupling} to approximate the lateral dynamics of the pelvis of Cassie during walking. Its dynamics are:
\begin{align}
 \ddot{y}_{\textrm{SSP}} &= \lambda^2 y,   \tag{SSP}  \\
 \ddot{y}_{\textrm{DSP}} &= 0,   \tag{DSP}
\end{align}
where $\lambda = \sqrt{ \frac{g}{z_0}} $, and $z_0$ is the averaged height of the point mass of aSLIP walking. The transition from SSP to DSP, $\Delta_{\textrm{S} \rightarrow \textrm{D}}$, and the transition from DSP to SSP,  $\Delta_{\textrm{D} \rightarrow \textrm{S}}$, are assumed to be smooth, thus the impact maps are defined as:
\begin{eqnarray}
\label{ImpactS2D}
 \Delta_{\textrm{S} \rightarrow \textrm{D}} &:& \left \{\begin{matrix}
\dot{y}^{+} = \dot{y}^{-}  \\
y^{+} = y^{-}
\end{matrix}\right.
\\
\label{ImpactD2S}
\Delta_{\textrm{D} \rightarrow \textrm{S}} &:& \left \{\begin{matrix*}[l]
\dot{y}^{+} = \dot{y}^{-}  \\
y^{+} = y^{-}   - w
\end{matrix*}\right.
\end{eqnarray}
where $w$ is the step width from the stance foot position to the landing foot. The transitions are time-based since $\{T_\textrm{SSP}, T_\textrm{DSP}\}$ are already determined by the aSLIP walking in the sagittal plane.

The periodic orbit for the lateral motion can be identified in closed-form from the linear dynamics of the H-LIP. For the lateral motion, we limit the choice to the period-2 (P2) orbits, which are periodic orbits with two steps. Fig. \ref{PeriodicWalking} (L2) illustrates different P2 orbits in the phase portrait. Note that the lines with slope $\sigma_2$ is called the lines of characteristics for the P2 orbits \cite{xiong2019IrosStepping}, where
\begin{equation}
\sigma_2 := \lambda {\textrm{tanh} \left(\frac{T_\textrm{SSP}}{2} \lambda \right)}.
\end{equation}
Any point of the line yields a periodic orbit by following the flow of the dynamics. More details can be found in \cite{xiong2019IrosStepping}. Fig. \ref{PeriodicWalking} (L3) shows the examples of several P2 orbits with the trajectories of the point mass.

\textbf{Orbit Composition.} Composition of the mass trajectory of the aSLIP in the sagittal plane and that of the H-LIP in the lateral plane yields the desired COM trajectory in 3D. The point foot nature of the two reduced order models locates the center of pressure at the center of the foot, which is expected to provide a maximum margin for foot stability for embedding the COM on the humanoid. Fig. \ref{PeriodicWalking} (C) shows an example of the composed trajectory in 3D.
\section{Transitional Motion Generation via Footed Reduced Order Models}
In this section, we describe \textit{footed} reduced order models for fast and dynamic motion transition to enable versatile walking behaviors. The foot actuation is directly applied on the reduced order models with the torque limits and the constraint that the center of pressure (COP) lies within the feet.

The transition behavior is defined as the motion that transits the robot states between standing configurations to periodic motions or between periodic motions themselves. Optimization on the transition motion of the full dimensional robot dynamics is still a difficult problem to solve due to complex specifications on the dynamics constraints, contact sequences, and actuation limits. The transition optimization on the footed reduced order models can simplify the optimization while providing the desired transition COM dynamics. The application of ankle actuation also provides fast and smooth transitions.

We assume that the transition motion is specifically realized within two domains, one DSP and one SSP. In other words, we pre-specify the contact sequence and number of domains for the transition motion. It is also possible to formulate the optimization into a contact implicit optimization using the Linear Complementarity Problem formulation \cite{posa2014direct} or using potential smooth techniques \cite{Tyler} for the hybrid dynamics, which is not within the scope of this paper.

\subsection{Footed aSLIP for Transition in Sagittal Plane}
The footed aSLIP model is the aSLIP with actuated foot (Fig. \ref{transition} (S1)). The foot actuation resembles the ankle pitch on the humanoid. The foot length equals to that of the humanoid. The dynamics are in the Appendix. The transition optimization formulation is similar to the previous aSLIP optimization. The differences are as follows.

\textbf{Cost Function.} With the foot actuation, the cost for the transitional optimization includes both the leg length actuation and the ankle torques:
\begin{equation}
 J =  \sum_{i = 1}^N \frac{\Delta t}{2} ({{\ddot{L}}_\text{L}}^{i^2} + {{\ddot{L}}_\text{R}}^{i^2} + {u^{i^2}_{\text{L}_y}} + {u^{i^2}_{\text{R}_y}}),
\end{equation}
where $u^i_{\text{L}_y}, u^i_{\text{R}_y}$ are the ankle pitch torques.

\textbf{ZMP and Actuation Constraints.} To ensure the optimized COM trajectories are realizable on the humanoid, the zero moment point (ZMP) and physical actuation constraints must be included,
\begin{eqnarray}
 -L_h F^i_{\text{L/R}}<  u^i_{{\text{L/R}}_y}< L_t  F^i_{\text{L/R}},  \
 \left |  u^i_{{\text{L/R}}_y} \right | < u_{y_\text{max}},
\end{eqnarray}
where $L_h, L_t$ are the distance between the projected ankle pitch axis to the heel and to the toe, $ F^i_{\text{L/R}}$ is the normal reaction force of foot contact, and $u_{y_\text{max}}$ is the maximum ankle pitch torque of the humanoid. The interpretation of the ZMP constraint is simply that the COP must be in the foot so that the foot does not rotate on the ground.

\textbf{Initial/Final States.} The initial and final states must be identical to the ones at the beginning and the end of the motion transition. The specification of the states comes from the humanoid. The real leg length $r$, leg angle $\beta$ and their velocities can directly mapped from the COM states of the robot. The leg length state can be solved by force balancing on the point mass with its acceleration. 

\subsection{Footed LIP for Transition via Quadratic Programs}
\begin{figure}[t]
      \centering
      \includegraphics[width= 3in]{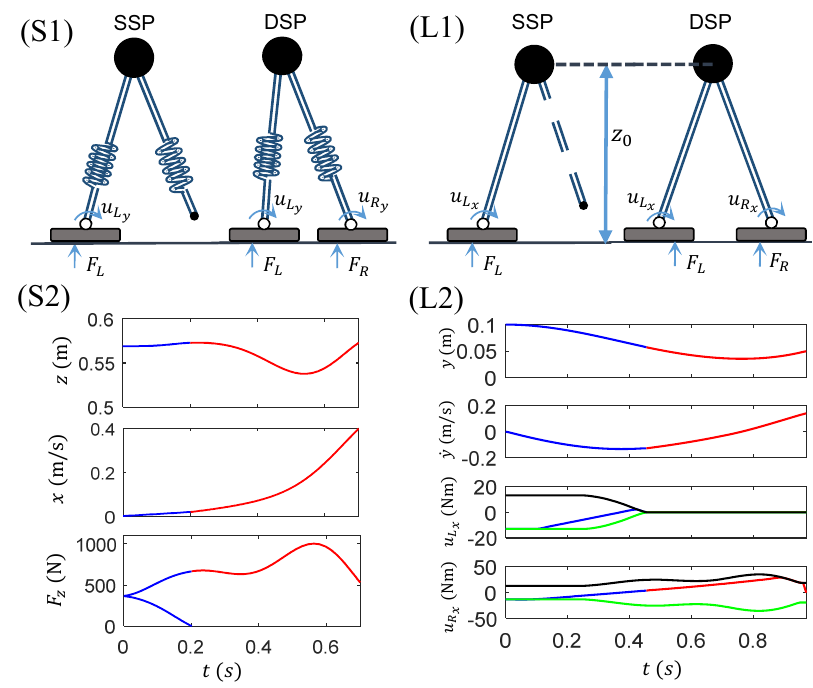}
      \caption{(S1, L1) Illustration of the footed aSLIP and footed H-LIP. (S2, L2) The results of optimizations of the footed reduced order models transiting from standing to periodic walking. The blue lines indicate those in DSP, and the red indicate those in SSP. The yellow and black indicate the lower and upper bounds respectively on the ZMP constraint in the H-LIP optimization.}
      \label{transition}
\end{figure}

The durations of each domain in the transitional motion are determined by the footed aSLIP optimization. For generating the corresponding motion in the lateral plane, we use the footed H-LIP model, which is the H-LIP with foot actuation (Fig. \ref{transition} (L1)). The ankle actuation on the foot becomes the input to the system. Its continuous dynamics can be written compactly in each domain as:
 \begin{align}
 \ddot{y}_{\textrm{SSP}} &= \lambda^2 y + \frac{1}{m z_0} u_{\text{L/R}_x},   \tag{SSP}  \\
 \ddot{y}_{\textrm{DSP}} &=  \frac{1}{m z_0}(u_{\text{L}_x} + u_{\text{R}_x}),   \tag{DSP}
\end{align}
where $u_{\text{L}_x}$ and $u_{\text{R}_x}$ are the ankle roll actuations. $u_{\text{L/R}_x} = 0$ if the foot is not in contact with the ground. $z_0$ is the averaged height of the mass of the footed aSLIP in transition motion. 

The linear dynamics motivates a quadratic program (QP) formulation for optimizing the transition from an initial state $[y_0; \dot{y}_0]$ to a final state $[y_f; \dot{y}_f]$. We discrete the trajectory over time with the discretized linear dynamics being,
 \begin{align}
\underset{Y^{k+1}}{\underbrace{\begin{bmatrix}
y^{k+1} \\
\dot{y}^{k+1}
\end{bmatrix}}}
=
A^{\text{DSP/SSP}}
\underset{Y^k}{\underbrace{\begin{bmatrix}
y^k \\
\dot{y}^k
\end{bmatrix} }}
 +
\underset{B}{\underbrace{\begin{bmatrix}
\frac{T^2}{2m z_0} \\
\frac{T}{m z_0}
\end{bmatrix} }} u^{\text{DSP/SSP}}_x,
\end{align}
where,
 \begin{align}
 A^{\text{SSP}} = \begin{bmatrix}
1 & T\\
\lambda T & 1
\end{bmatrix},  \ \  A^{\text{DSP}} = \begin{bmatrix}
1 & T \\
0 & 1
\end{bmatrix},\\
 u^{\text{SSP}} = u_{\text{L/R}_x}, \ \  u^{\text{DSP}} =u_{\text{L}_x} + u_{\text{R}_x}.
 \end{align}

\textbf{ZMP and Actuation Constraints.} With an eye towards the embedding on the humanoid, we also enforce the ZMP and actuation constraints in the lateral transition as:
\begin{equation}
\label{lateralZMP}
 -W_1 F^k_{\text{L/R}}   \leq  u^k_{\text{L/R}_x}  \leq   W_2 F^k_{\text{L/R}},  \
 \left |  u^k_{\text{L/R}_x} \right | < u_{x_\text{max}},
\end{equation}
where $W_1, W_2$ are the distances between the projected ankle roll axis to each edge of the foot, and $u_{x_\text{max}}$ is the maximum ankle roll torque.

\textbf{Coupling with Footed aSLIP.} The footed ROMs in each plane are coupled by the ground normal forces. The vertical reaction forces in Eq. \eqref{lateralZMP} are from $F_{\text{L/R}}(t)$ of the footed aSLIP\footnotemark.

\footnotetext[1]{The sum of the vertical reaction forces on the H-LIP, i.e. $F_L(t) + F_R(t)$, does not equal to $mg$, since the COM height varies from the footed aSLIP.}

\textbf{Quadratic Program.} The final QP formulation with minimizing $u$ for efficiency is as follows,
\begin{align}
\label{QP-LIP}
Y_{\{1,..., N \}}= & \underset{ \{u_{\text{L}_x}, u_{\text{R}_x} , Y \} \in \mathbb{R}^{N \times N \times 2N}    }
 {\text{argmin}}   \sum\nolimits_{k=1}^N u_{\text{L}_x}^{k^2} + u_{\text{R}_x}^{k^2} \\
\text{s.t.} \quad &  Y^{k+1} = A^{\text{SSP/DSP}} Y^k + B u^{\text{SSP/DSP}}_x  \: \ \ \quad\text{(H-LIP)}  \nonumber\\
\quad &  \left |  u^k_{\text{L/R}_x} \right | < u^x_{\text{max}}  \quad \ \ \quad\quad\quad\quad\quad \text{(Torque Limit)}  \nonumber \\
\quad &  -W_1 F^k_{\text{L/R}}   \leq  u^k_{\text{L/R}_x}  \leq   W_2 F^k_{\text{L/R}}  \quad \ \ \quad\quad\text{(ZMP)}  \nonumber \\
 \quad &  [y_N; \dot{y}_N]= [y_f; \dot{y}_f] \quad \ \ \quad \quad \quad\quad \, \text{(Final State)}   \nonumber \\
  \quad &  [y_1; \dot{y}_1]= [y_0; \dot{y}_0]. \quad \quad  \quad\quad\quad\quad \text{(Initial State)}   \nonumber
\end{align}

\textbf{Comparison with ZMP Approaches.} Quadratic programs \cite{kajita2003biped} \cite{brasseur2015robust} have been widely used on the LIP dynamics for controlling humanoid walking. In our approach, planning on the ZMP trajectory is not required and the QP is only required to be solved once instead of online recursively.

\textbf{Remark.} The optimizations on the footed reduced order models are connected by the ground reaction force. Therefore the footed aSLIP optimization and the footed H-LIP QP can be combined into a single nonlinear optimization. The steps are straightforward by combining the variables, constraints and cost functions. 

\section{Dynamics Embedding on Humanoids}
With the constructed COM dynamics from reduced order models, one can embed the dynamics on the fully actuated humanoid robot by tracking the desired COM trajectory. The walking is thus realized by defining all the required outputs and zeroing the outputs via feedback control. In this section, we briefly define the outputs for walking of the Exo and then describe the feedback control with ground reaction force matching using the control Lyapunov function based Quadratic programs (CLF-QPs).

\subsection{Output Definition}
\subsubsection{Outputs for DSP}
In DSP, foot contacts introduce 12 holonomic constraints. There are 6 outputs required for the 18 DOF robot. Except for the COM position, we require the pelvis orientation to be fixed. Thus the outputs for the DSP is defined as,
\begin{equation}
 \mathcal{Y}_{\textrm{DSP}}(q,t) = \begin{bmatrix}
  \mathbf{p}_{\textrm{COM}}(q) \\
    \pmb{\phi}_{\textrm{pelvis}}(q)
     \end{bmatrix} -
\begin{bmatrix}
 \mathbf{p}^{\text{desired}}_{\textrm{COM}}(t) \\
 \mathbf{0}  \end{bmatrix}.
\end{equation}

\subsubsection{Outputs for SSP}
In SSP, only one foot contacts the ground. Thus we define 12 outputs with additional 6 being on the swing foot position and orientation,
\begin{equation}
 \mathcal{Y}_{\textrm{SSP}}(q,t) = \begin{bmatrix}
\mathbf{p}_{\textrm{COM}}(q) \\
  \pmb{\phi}_{\textrm{pelvis}}(q)  \\
    \mathbf{p}_{\textrm{swingFoot}}(q) \\
     \pmb{\phi}_{\textrm{swingFoot}}(q)   \end{bmatrix} -
\begin{bmatrix}
 \mathbf{p}^{\text{desired}}_{\textrm{COM}}(t) \\
 \mathbf{0}\\
   \mathbf{p}^{\text{desired}}_{\textrm{swingFoot}}(t) \\
 \mathbf{0}   \end{bmatrix}.
\end{equation}
The swing foot position $\mathbf{p}^{\text{desired}}_{\textrm{swingFoot}}(t)$ is constructed smoothly from the initial to the final position of the SSP.
\begin{figure}[b]
      \centering
      \includegraphics[width= 2.7in]{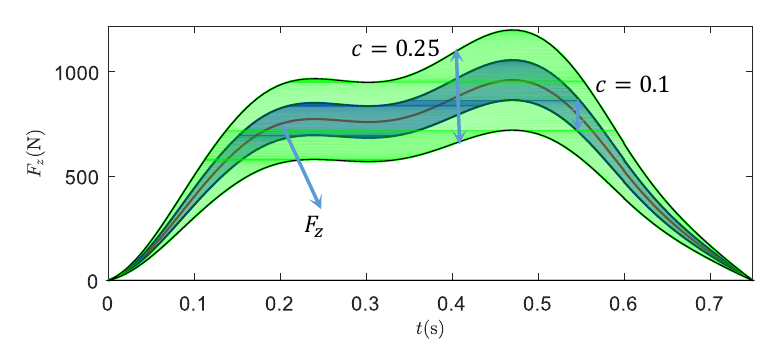}
      \caption{Illustration of the valid range of the relaxation on the desired normal force with different relaxation coefficient $c$.}
      \label{relaxation}
\end{figure}
\begin{figure*}[!t]
      \centering
      \includegraphics[width = 6.95in]{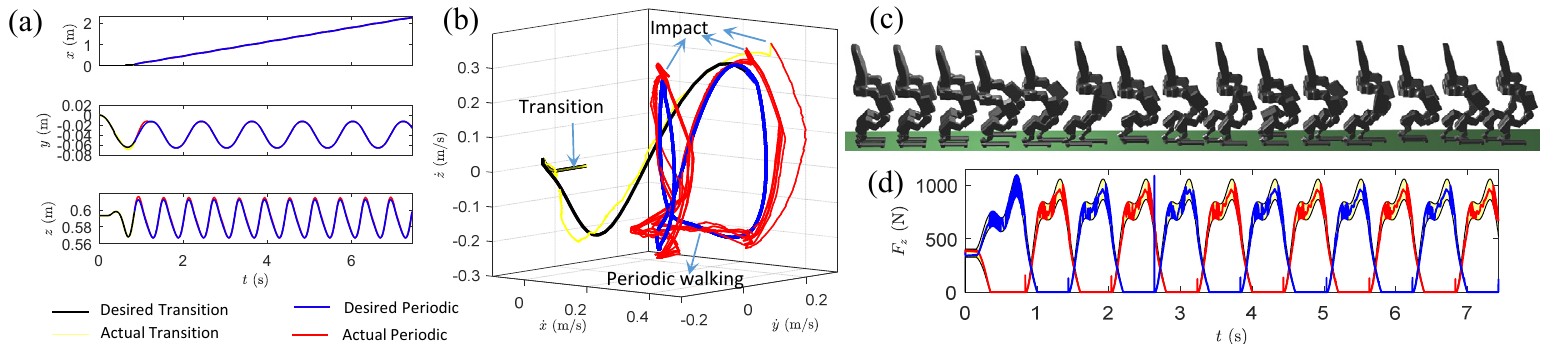}
      \caption{Simulation results of the walking from standing. (a) The tracking on the desired COM position. (b) The evolution of COM velocity. (c) Snopshots of the walking of the Exo. (d) Ground normal reaction forces are with in the relaxation under the CLF-QP control. }
      \label{results}
\end{figure*}
\subsection{CLF-QP}
We apply the control Lyapunov function based Quadratic programs (CLF-QP) \cite{ames2013towards} \cite{xiong2018bipedal} for zeroing the outputs. The control Lyapunov function $V$ is constructed over the feedback linearized output dynamics by quadratic functions of the outputs. The exponential convergence of $V$ is nicely enforced by the inequality condition on the derivative $\dot{V}$,
\begin{equation}
\dot{V}(u, q, \dot{q}) \leq - \gamma V(q, \dot{q}),
\end{equation}
with $\gamma >0$. The inequality is affine with respect to $u$,
\begin{equation}
A_v^{\textrm{CLF}}(q,\dot{q}) u \leq b_v^{\textrm{CLF}}(q,\dot{q}).
\end{equation}
Thus quadratic programs can be formulated for optimizing $u$ subject to this Lyapunov inequality and additional phyical constraint such as torque limits and contact constraints. Details about constructing the CLF-QPs are in \cite{ames2013towards} \cite{xiong2018bipedal}.

For the application of walking system with holonomic constraints, \cite{ames2013towards} \cite{xiong2018bipedal} also include the contact forces as the optimization variables. Holonomic constraints and ground contact constraints are thus encoded via equality and inequality constraints, respectively. Here we remove them from the optimization variables for QP formulation with minimum arguments for efficiency. It can be done since the contact forces are affine functions with respective to $u$,
\begin{align}
 F_v &= A_v u + b_v, \label{utoF}\\
 A_{v} &=  - (J_v M^{-1} J_v^{T} )^{-1} J_v M^{-1} B,\\
 b_{v} &= (J_v M^{-1} J_v^{T} )^{-1} (J_v M^{-1}h -\dot{J}_v \dot{q} ),
\end{align}
where $F_v$ is the contact force vector in domain $v$. It is derived from Eq. \eqref{eom} \eqref{hol}. The constraints on the ground reaction forces (GRF) are thus enforced directly on $u$. Let $C_v$ denote the contact constraint matrix, i.e. $C_v F_v \leq 0$, which includes the constraints of the friction cones, nonnegative normal forces and ZMP \cite{grizzle2014models}. Then the GRF constraint becomes,
\begin{equation}
C_v A_v u \leq -C_v b_v.
\end{equation}

\subsection{Force Control Embedding}
We would like the realized walking to exhibit the aSLIP walking behavior in terms of ground reaction forces. If the contact forces behave identically to those of the aSLIP walking, the transitions from DSP to SSP can happen naturally as the normal force goes to zero. The transition from SSP to DSP also behaves smoothly when the normal force increases matching the aSLIP. Thus the force control can be realized via enforcing $ S_v F_v = F^z_{\text{aSLIP}}$, where $S_v$ is the selection matrix. With Eq. \eqref{utoF}, the equality becomes,
\begin{equation}
\label{ForceEquality}
S_v A_v u = F^z_{\text{aSLIP}}  -S_v b_v.
\end{equation}
Since we control the vertical COM oscillation of the humanoid to be identical to that of the aSLIP simultaneously, the vertical GRF match needs to be relaxed. Otherwise, strictly matching the reaction force can destabilize the tracking. Thus we relax the equality in Eq.\eqref{ForceEquality} by the inequality $\left |  S_v F_v  - F^z_{\text{aSLIP}} \right |   \leq  c F^z_{\text{aSLIP}}$, resulting in:
\begin{equation}
\label{ForceRelaxation}
  \underset{c_{lb}}{\underbrace{(1-c) F^z_{\text{aSLIP}}- S_v b_v}} \leq  S_v A_v u   \leq \underset{c_{ub}}{\underbrace{ (1+c) F^z_{\text{aSLIP}} - S_v b_v}},
\end{equation}
where $c\in (0,1)$ is a coefficient of the relaxation. In practice, we use $c=0.1$. Fig. \ref{relaxation} shows the valid range of the $F_z$ with different $c$. It is important to note that this relaxation shrinks as the force decreases, which enforces the smooth transitions between DSP and SSP with the GRF decreasing to 0 or increasing from 0.

\subsection{Main Control Law}
The main feedback control loop can thus be realized by the CLF-QP with force control embedded as an inequality constraint:
\begin{align}
\label{QP}
\quad u^{*} =  \underset{u \in \mathbb{R}^{12}, \delta\in \mathbb{R}} {\text{argmin}}& u^T  H u + 2 F u + p\delta^2, \\
\text{s.t.} \quad  &  A_v^{\textrm{CLF}}(q,\dot{q}) u \leq b_v^{\textrm{CLF}}(q,\dot{q})  + \delta, \: \quad \ \quad  \text{(CLF)}\nonumber \\
 \quad &  C_v A_v u \leq -C_v b_v,   \ \  \quad  \quad  \quad  \quad \quad \quad \ \text{(GRF)}    \nonumber  \\
 \quad  & u_{lb} \leq u \leq u_{ub},  \nonumber  \quad \quad \quad  \quad   \quad   \text{(Torque Limit)}\\
 \quad  &  c_{lb} \leq   S_v A_v u \leq c_{ub}, \ \ \quad  \quad \text{(Force Control)}\nonumber
\end{align}
where $\delta$ is a relaxation term for increasing the instantaneous feasibility of the QP, and $p$ is a large positive penalty constant. Adding $\delta$ is relaxing the tracking performance. Since there is no guarantee for the feasibility of the QP with multiple constraints, it is appropriate to relax the performance while keeping the system safe.

\textbf{Remark.} In the context of QP based controllers, task/operational space control (TSC) \cite{wensing2013high} could be utilized in an similar fashion. The CLF-QP encodes the control objective by an inequality constraint; while the TSC encodes the objective in the cost function, which minimizes the difference between the actual output accelerations and the desired values from a PD control. The ground reaction force matching can also be put into the TSC formulation.

%

\section{Results}
The proposed framework is primarily implemented in simulation. The dynamics are numerically integrated using MATLAB's ode45 function with event functions for triggering different contact domains. The nonlinear programs for the trajectory optimization of the aSLIP are solved by IPOPT \cite{wachter2006implementation}. The QP for the footed H-LIP and the CLF-QP for the control of the Exo are solved by qpOASES \cite{Ferreau2014} with active set method. We mainly evaluate the method for two walking scenarios, i.e. periodic walking from standing and transition between two periodic walking behaviors.

For both cases, periodic walking is first composed from the periodic orbits of the aSLIP and the H-LIP. Then the nonlinear program on the footed aSLIP and the QP for the footed H-LIP are solved for generating the transition in each plane. The ground normal forces of the aSLIP are used as references in Eq. \eqref{ForceRelaxation}. The composed trajectories of the COM are set to the desired outputs for the CLF-QP to generate desired torques in the control loop.

Fig. \ref{results} shows one of the simulation results. Additional simulation and video can be found in \cite{Supplementary}. The COM of the Exo follows the desired COM trajectory well. The ground normal forces are within the range of the relaxation. Note that the impact of the foot-ground contact still exists in the walking since we do not necessarily require the foot to strike the ground with zero velocity. It does not destabilize the system due to the force control. On the hardware, we expect in future work to use the load cells for contact detection and use the force control to generate compliant behavior instantaneously after impact.

\section{Conclusion and Future Work}
In this paper, we propose the decoupling of walking in terms of transitional and periodic walking, each of which is further decoupled into walking in the sagittal and lateral plane. Optimizations for planar reduced order models (ROMs) are utilized for motion generation and 3D composition. Desired walking behavior is thus encoded by the COM dynamics and the ground reaction forces of the ROMs. Control Lyapunov function based quadratic programs (CLF-QPs) with active force control are applied for the dynamics and force embedding on the fully actuated humanoid robots.

Future work will be heavily focused toward the hardware implementation of the proposed approach on the Exoskeleton. This will include online gait generation via the reduced order models and realization of the CLF-QP with contact force control.
\addtolength{\textheight}{-0.0cm}

\section*{APPENDIX}
\subsubsection{Dynamics of the Point Foot and Footed aSLIP}
The dynamics the footed aSLIP is,
\begin{align}
\textrm{S}:&\left\{\begin{matrix*}[l]
\ddot{r}_1 = \tfrac{F_1}{m} - g \text{cos}(\beta_1) + r \dot{\beta}_1^2
\\ \ddot{\beta}_1 =\tfrac{1}{r_1} ( -2\dot{\beta_1} \dot{r}_1 + g \text{sin}(\beta_1) + \frac{u_1}{m r_1} )
\\  \ddot{s}_1 = \ddot{L}_1 - \ddot{r}_1
\end{matrix*}\right., \nonumber \\
\textrm{D}:&\left\{\begin{matrix*}[l]
\ddot{r}_1 = \frac{F_1 + F_2 \text{cos}(\delta_q)}{m} - g \text{cos}(q_1) + r_1 \dot{q}_1^2  + \frac{u_2}{m r_2} \text{sin}(\delta_q) \\
 \ddot{q}_1 =\frac{ -2\dot{q}_1 \dot{r}_1 + g \text{sin}(q_1) - \frac{ F_2}{m} \text{sin}(\delta_q) }{r_1}   + \frac{\text{cos}(\delta_q)}{m r_1 r_2}u_2 + \frac{u_1}{m r_1^2}\\
  \ddot{r}_2 = \frac{F_2 + F_1 \text{cos}(\delta_q)}{m} - g \text{cos}(q_2) + r_2 \dot{q}_2^2  -  \frac{u_1}{m r_1}\text{sin}(\delta_q)\\
   \ddot{q}_2 =\frac{-2\dot{q}_2 \dot{r}_2 + g \text{sin}(q_2) + \frac{ F_1}{m} \text{sin}(\delta_q) }{r_2} + \frac{\text{cos}(\delta_q)}{m r_1 r_2}u_1 + \frac{u_2}{m r_2^2} \\
   \ddot{s}_1 = \ddot{L}_1 - \ddot{r}_1 \\
    \ddot{s}_2 = \ddot{L}_2 - \ddot{r}_2
\end{matrix*}\right.,\nonumber
\end{align}
where $\delta_q = q_1 -q_2$,  $\ddot{L}_{1/2}$ are assumed to be the virtual input \cite{xiong2018coupling}, and $u_{1/2}$ are the ankle actuation. The spring forces $F_{1/2}$ come from the leg spring \cite{xiong2018bipedal}. The dynamics of point foot aSLIP is that of the footed aSLIP with zero ankle actuation.

\subsubsection{Impact Map Assumption} The impact map is,
\begin{eqnarray}
\Delta_{\textrm{SSP} \rightarrow \textrm{DSP}}:\left\{\begin{matrix*}[l]
 \dot{q}^{+}_2 = \frac{1}{r_2}( \dot{q}_1 \dot{r}_1 \text{cos} (\delta q) + \dot{r}_1 \text{sin} (\delta q) )\\
 \dot{r}^{+}_2 = \dot{r}_1 \text{cos} (\delta q) -  \dot{q}_1 r_1  \text{sin} (\delta q)\\
 \dot{s}^{+}_2 = \dot{L}^{-}_2 - \dot{r}^{+}_2
 \end{matrix*}\right. , \nonumber
\end{eqnarray}
where we only specify the discontinuous states to save space. The transition map $\Delta_{\textrm{DSP} \rightarrow \textrm{SSP}}$ is smooth.
\subsubsection{Leg Spring Approximation for Cassie}
The nonlinear prismatic spring on the aSLIP model comes from the approximation of the leg dynamics of Cassie \cite{xiong2018bipedal}. Here we detail the coefficients for reference.
\begin{eqnarray}
\begin{matrix*}[l] K(L)=(23309 L^4-  55230 L^2+ 48657 L-9451  )c,\\
 D(L) = (348 L^4 - 824 L^2+ 726 L-141)c, \end{matrix*} \nonumber
\end{eqnarray}
where $c = 2.28$ is the mass ratio between the humanoids and Cassie, and the leg spring force is $F =  K(L)s + D(L) \dot{s}$,
\bibliographystyle{IEEEtran}
\bibliography{Walking-exo}
\end{document}